\documentclass[letterpaper,conference,compsoc]{IEEEtran}

\usepackage[english]{babel}

\usepackage{graphics} 		
\usepackage{graphicx}
\usepackage{epsfig} 		
\usepackage{epstopdf}

\usepackage{listings}
\usepackage{color}
\usepackage{nameref}
\usepackage[bookmarks=false]{hyperref}
\usepackage[colorinlistoftodos]{todonotes}
\usepackage{index}
\usepackage{amsmath}	 	
\usepackage{amssymb}  		

\usepackage{dsfont}
\usepackage{mathtools}

\usepackage{epigraph}
\usepackage{lscape}
\usepackage[]{nomencl}		
\usepackage[ruled,vlined]{algorithm2e}
\usepackage{multicol}
\usepackage{multirow}
\usepackage{etoolbox}

\usepackage{caption}
\usepackage{subcaption}
\usepackage{wrapfig}

\usepackage{units}
\usepackage[round]{natbib}

\DeclareMathOperator{\diag}{\mathrm{diag}} 

\newcommand{\fig}{Figure~}
\newcommand{\eq}{Equation~}

\newcommand{\tab}{Table~}

\definecolor{darkgreen}{rgb}{0,0.5,0}		
\definecolor{purple}{rgb}{0.75,0,0.75}
\definecolor{pink}{rgb}{1,0.4,0.6}

\definecolor{matlab1}{rgb}{0,0,1}
\definecolor{matlab2}{rgb}{0,0.5,0}
\definecolor{matlab3}{rgb}{1,0,0}
\definecolor{matlab4}{rgb}{0,0.75,0.75}
\definecolor{matlab5}{rgb}{0.75,0,0.75}
\definecolor{matlab6}{rgb}{0.75,0.75,0}
\definecolor{matlab7}{rgb}{0.25,0.25,0.25}

\renewcommand{\vec}{\boldsymbol}	
\newcommand{\mat}{\boldsymbol}		
\newcommand{\R}[0]{\mathds{R}}		

\newcommand{\inv}[0]{^{-1}} 		
\newcommand{\T}[0]{^T} 				

\newcommand{\asin}[0]{\sin^{-1}} 	%

\newcommand{\prob}{{p}} 
\newcommand{\gauss}[2]{\mathcal{N} \big( #1,#2 \big) }	

\newcommand{\D}[0]{\mathds{D}} 

\newcommand{\GP}[0]{\mathcal{GP}} 	
\newcommand{\coveq}[0]{k} 

\newcommand{\parameters}[0]{\vec \theta}

\newcommand{\nntfNo}[0]{\sigma} 	
\newcommand{\nntf}[1]{\nntfNo \left( #1 \right)} 	
\newcommand{\nnB}[0]{\vec B} 				
\newcommand{\nnW}[0]{\mat W} 				

\newcommand{\inputs}{\mat X}
\newcommand{\targets}{\mat Y}

\newcommand{\latents}{\mat H}

\newcommand{\inputsSpace}{\mathcal{X}}
\newcommand{\targetsSpace}{\mathcal{Y}}
\newcommand{\latentsSpace}{\mathcal{H}}
\newcommand{\reallatentsSpace}{\mathcal{L}}

\newcommand{\dimInputs}{D}
\newcommand{\dimLatents}{Q}

\newcommand{\nntransformation}[0]{\mathds{T}}

\newcommand{\newmethodlong}[0]{Manifold Gaussian Processes} 	%
\newcommand{\newmethodshort}[0]{mGP} 	

\newcommand{\mappingNo}[0]{M}
\newcommand{\regressionNo}[0]{F}
\newcommand{\gprNo}[0]{G}
\newcommand{\gpkernel}[0]{covariance function}

\newcommand{\mapping}[1]{{\mappingNo \hspace{-2pt} \left( #1 \right) }}
\newcommand{\regression}[1]{{\regressionNo \hspace{-2pt} \left( #1 \right) }}

\newcommand{\secExpOne}[0]{Step Function}
\newcommand{\secExpTwo}[0]{Bipedal Robot Locomotion}

\newcommand{\secExpFour}[0]{Multiple Length-Scales}

\newcommand{\sizeLayer}[0]{q}

\DeclareCaptionType{copyrightbox}

\begin{document}

\title{Manifold Gaussian Processes for Regression}

\author{\IEEEauthorblockN{Roberto Calandra\IEEEauthorrefmark{1}, Jan Peters\IEEEauthorrefmark{1}\IEEEauthorrefmark{2}, Carl Edward Rasmussen\IEEEauthorrefmark{3} and Marc Peter Deisenroth\IEEEauthorrefmark{4}}

\IEEEauthorblockA{\IEEEauthorrefmark{1}Intelligent Autonomous Systems Lab, Technische Universit\"at Darmstadt, Germany}
\IEEEauthorblockA{\IEEEauthorrefmark{2}Max Planck Institute for Intelligent Systems, T\"ubingen, Germany}
\IEEEauthorblockA{\IEEEauthorrefmark{3}Department of Engineering, University of Cambridge, United Kingdom}
\IEEEauthorblockA{\IEEEauthorrefmark{4}Department of Computing, Imperial College London, United Kingdom}
}

\maketitle


\begin{abstract}
%
Off-the-shelf Gaussian Process (GP) covariance functions encode smoothness assumptions on the structure of the function to be modeled. 
To model complex and non-differentiable functions, these smoothness assumptions are often too restrictive.
%
%
One way to alleviate this limitation is to find a different representation of the data by introducing a feature space. 
This feature space is often learned in an unsupervised way, which might lead to data representations that are not useful for the overall regression task.
%
%
In this paper, we propose Manifold Gaussian Processes, a novel supervised method that jointly learns a transformation of the data into a feature space and a GP regression from the feature space to observed space. 
The Manifold GP is a full GP and allows to learn data representations, which are useful for the overall regression task.
%
%
As a proof-of-concept, we evaluate our approach on complex non-smooth functions where standard GPs perform poorly, such as step functions and robotics tasks with contacts.%
%
\end{abstract}

\IEEEpeerreviewmaketitle


\section{Introduction}
\label{Introduction}



Gaussian Processes (GPs) are a powerful state-of-the-art nonparametric
Bayesian regression method.
The {\gpkernel} of a GP implicitly encodes high-level assumptions about the
underlying function to be modeled, e.g., smoothness or periodicity. Hence, the choice of a suitable
{\gpkernel} for a specific data set is crucial. A standard choice is the squared exponential (Gaussian) {\gpkernel}, which implies assumptions, such as smoothness and stationarity.
Although the squared exponential can be applied to a great range of problems, generic {\gpkernel}s may also be inadequate to model a variety of
functions where the common smoothness assumptions are violated, such
as ground contacts in robot locomotion.

Two common approaches can overcome the limitations of standard {\gpkernel}s.
The first approach combines multiple standard {\gpkernel}s to form a
new {\gpkernel}~\citep{Rasmussen2006,Wilson2013,Duvenaud2013a}. 
This approach allows to automatically design relatively complex
{\gpkernel}s. However, the resulting {\gpkernel} is still limited by
the properties of the combined {\gpkernel}s.
%
The second approach is based on data transformation (or
pre-processing), after which the data can be modeled with standard
{\gpkernel}s. 
One way to implement this second approach is to transform the output
space as in the Warped GP~\citep{Snelson2004}. An alternative is to
transform the input space.
Transforming the input space and subsequently applying GP regression
with a standard {\gpkernel} is equivalent to
GP regression with a new {\gpkernel} that explicitly depends on the
transformation~\citep{MacKay1998}. 
One example is
the stationary periodic {\gpkernel}~\citep{MacKay1998,HajiGhassemi2014}, which
effectively is the squared exponential {\gpkernel} applied to a
complex representation of the input variables.
Common transformations of the inputs include data normalization and
dimensionality reduction, e.g., PCA~\citep{Pearson1901}. Generally,
these input transformations are good heuristics or optimize an
unsupervised objective. However, they may be suboptimal for the overall
regression
task. 

%
In this paper, we propose the \emph{Manifold Gaussian Process}
({\newmethodshort}), which is based on MacKay's ideas to devise flexible
covariance functions for GPs. Our GP model is equivalent to jointly
learning a data transformation into a feature space followed by a GP
regression with off-the-shelf covariance functions from feature space
to observed space. The model profits from standard GP properties, such
as a straightforward incorporation of a prior mean function and a faithful
representation of model uncertainty.

  %
      Multiple related approaches in the literature attempt joint
        supervised learning of features and regression/classification.
	In~\citet{Salakhutdinov2007a}, pre-training of the input
        transformation makes use of computationally expensive
        unsupervised learning that requires thousands of data
        points. 
        \citet{Snoek2012b} combined both unsupervised and supervised objectives for the
        optimization of an input transformation in a classification task.
\begin{figure*}[th]
		\centering

		\begin{subfigure}[t]{0.4\hsize}
			\includegraphics[width=\textwidth]{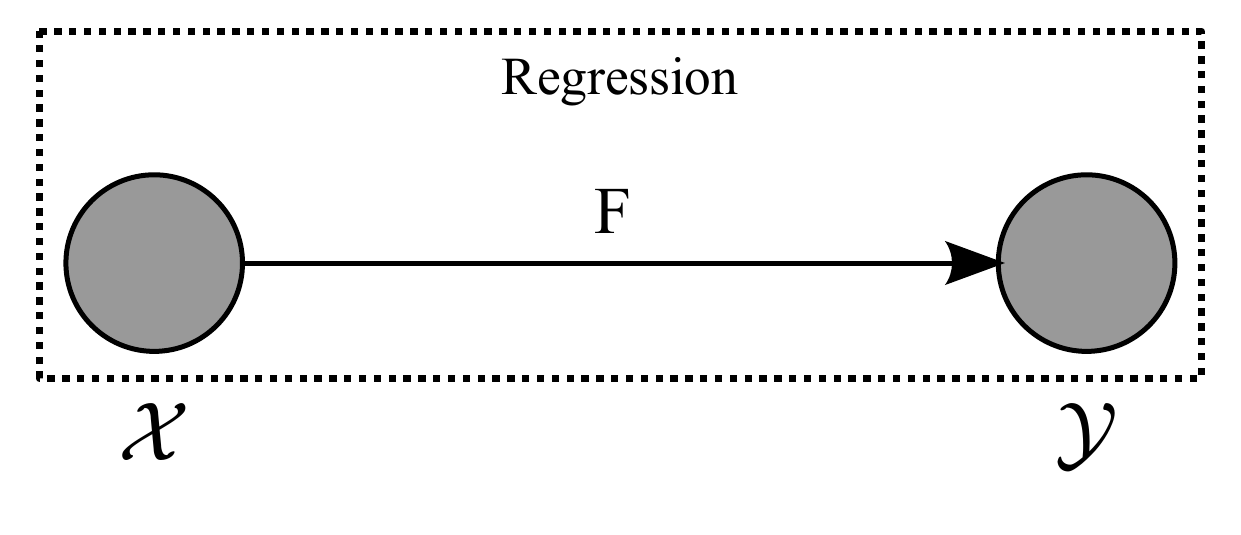}
			\caption{Supervised learning of a regression function.}
			\label{fig:model_1}
		\end{subfigure}%
		\hspace{5mm}
		\begin{subfigure}[t]{0.4\hsize}
			\includegraphics[width=\textwidth]{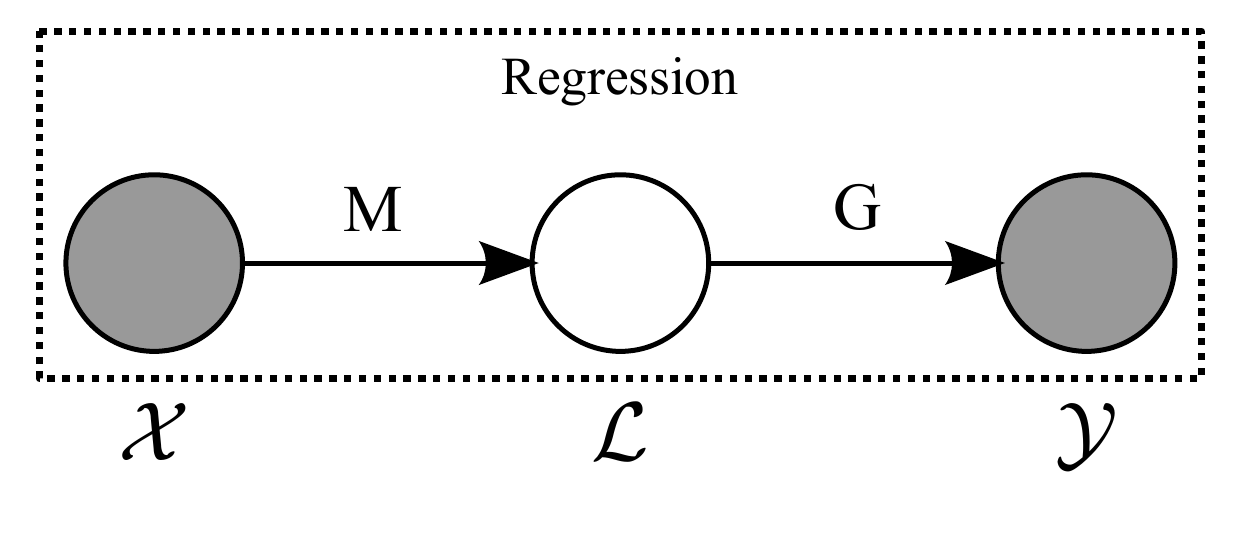}
			\caption{Supervised learning integrating out a latent space is \emph{intractable}.}
			\label{fig:model_2}
		\end{subfigure}
			\\
		\begin{subfigure}[t]{0.4\hsize}
			\includegraphics[width=\textwidth]{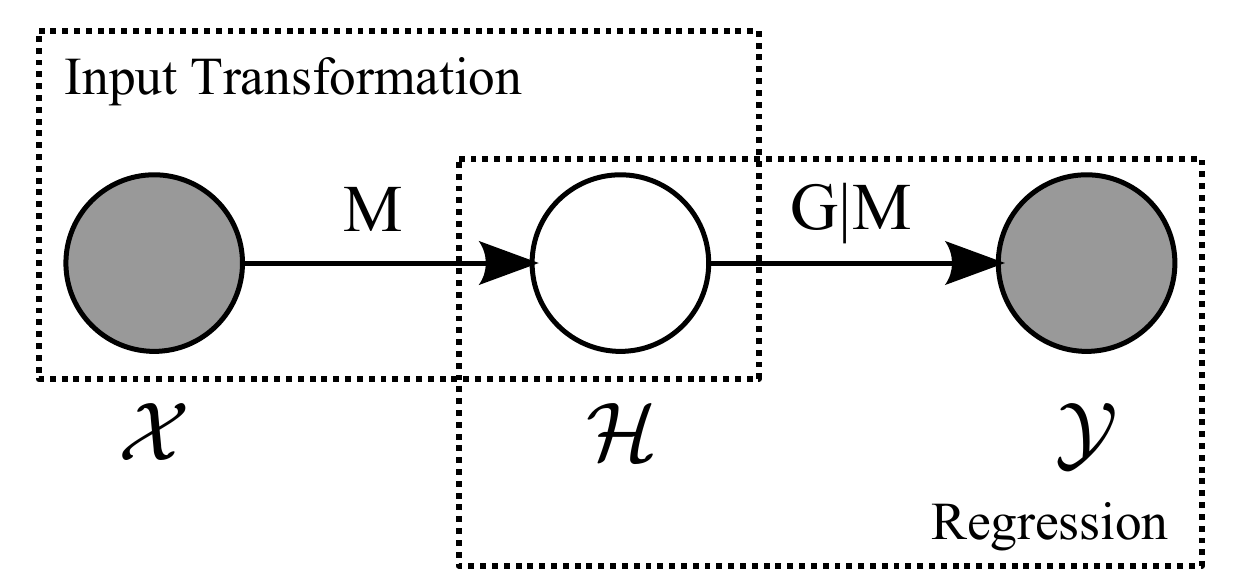}
			\caption{Unsupervisedly learned input transformation~$\mappingNo$ followed by a conditional regression~$\gprNo|\mappingNo$.}
			\label{fig:model_3}
		\end{subfigure}
		\hspace{5mm}
		\begin{subfigure}[t]{0.4\hsize}
			\includegraphics[width=\textwidth]{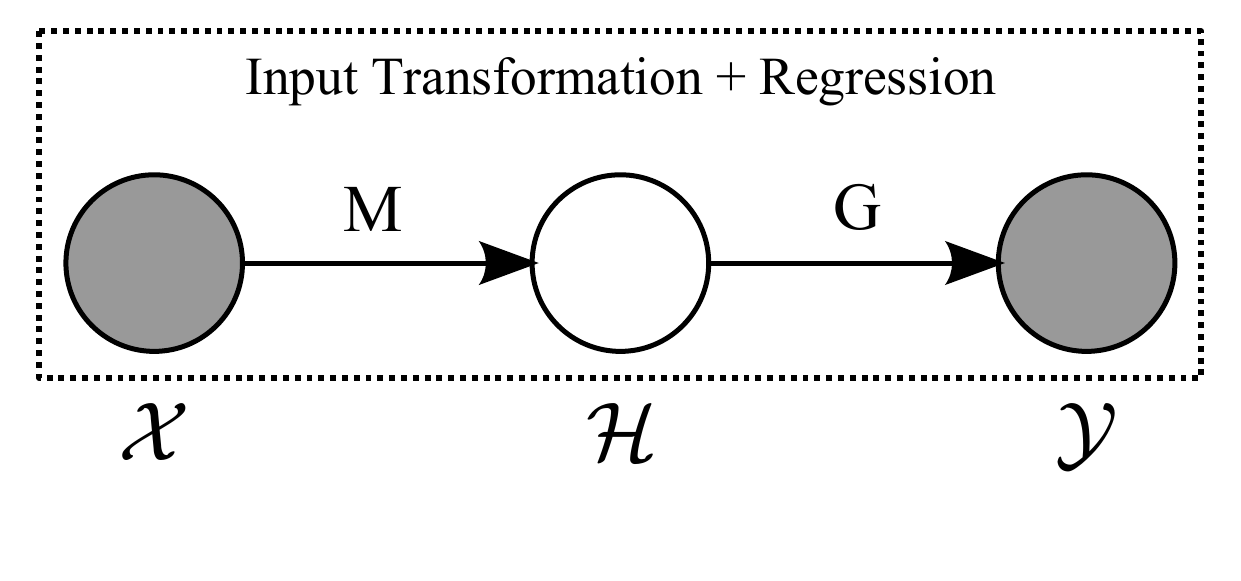}
			\caption{Manifold GP: joint supervised learning of the input transformation~$\mappingNo$ and the regression task~$\gprNo$.}
			\label{fig:model_4}
		\end{subfigure}
		\caption{Different regression settings to learn the function $\regressionNo: \inputsSpace
                  \to \targetsSpace$. (\subref{fig:model_1})
                  Standard supervised regression.
                  (\subref{fig:model_2}) Regression with an auxiliary
                  latent space~$\reallatentsSpace$ that allows to
                  simplify the task. In a full Bayesian framework,
                  $\reallatentsSpace$ would be integrated out, which
                  is analytically intractable.  (\subref{fig:model_3})
                  Decomposition of the overall regression
                  task~$\regressionNo$ into discovering a feature
                  space~$\latentsSpace$ using the map~$\mappingNo$ and
                  a subsequent (conditional)
                  regression~$\gprNo|\mappingNo$. (\subref{fig:model_4})
                  Our {\newmethodshort} learns the mappings $\gprNo$
                  and $\mappingNo$ \emph{jointly}.}
		\label{fig:model}
	\end{figure*}
	%
%
        Unlike these approaches, the mGP is motivated by the need of a
        stronger (i.e., supervised) guidance to discover suitable
        transformations for regression problems, while remaining
        within a Bayesian framework.
        \citet{Damianou2013a} proposed the Deep GP, which stacks multiple layers of GP-LVMs, similarly to a neural network. This model exhibits great flexibility in supervised and unsupervised settings, but the resulting model is not a full GP.
        \citet{Snelson2006} proposed a supervised dimensionality reduction by jointly learning a liner transformation of the input and a GP.
        \citet{Snoek2014a} transformed the input data using a Beta distribution whose parameters were learned jointly with the GP.
        However, the purpose of this transformation is to account for skewness in the data, while {\newmethodshort} allows for a more general class of transformations.



\section{Manifold Gaussian Processes}
\label{Method}

In the following, we review methods for regression, which may use
latent or feature spaces. Then, we provide a brief introduction to
Gaussian Process regression. Finally, we introduce the
{\newmethodlong}, our novel approach to jointly learning a regression
model and a suitable feature representation of the data.
%
%
\subsection{Regression with Learned Features}

        We assume $N$ training inputs $\vec x_n \in \inputsSpace
        \subseteq \R^{\dimInputs}$ and respective outputs $y_n \in
        \targetsSpace \subseteq \R$, where $y_n = \regressionNo(\vec
        x_n) + w$, $w \sim \gauss{0}{\sigma_w^2}$, $n = 1, \dots,
        N$. The training data is denoted by $\inputs$ and $\targets$
        for the inputs and targets, respectively. We consider
        the task of learning a regression function
        \mbox{$\regressionNo: \inputsSpace \to \targetsSpace$}.
	The corresponding setting is given in \fig\ref{fig:model_1}.
	Discovering the regression function~$\regressionNo$ is often
        challenging for
	nonlinear functions.
	A typical way to simplify and distribute the complexity of the
        regression problem is to introduce an auxiliary latent
        space~$\reallatentsSpace$. The function $\regressionNo$ can
        then be decomposed into \mbox{$\regressionNo = \gprNo \circ
          \mappingNo$}, where \mbox{$\mappingNo: \inputsSpace \to
          \reallatentsSpace$} and $\gprNo: \reallatentsSpace \to
        \targetsSpace$, as shown in \fig\ref{fig:model_2}.
	In a full Bayesian framework, the latent
        space~$\reallatentsSpace$ is integrated out to solve the
        regression task~$\regressionNo$, which is often analytically
        unfeasible~\citep{Schmidt2003}.

	A common approximation to the full Bayesian framework is to
        introduce a deterministic feature space~$\latentsSpace$, and
        to find the mappings $\mappingNo$ and $\gprNo$ in two
        consecutive steps. First, $\mappingNo$ is determined by means
        of unsupervised feature learning. Second, the regression
        $\gprNo$ is learned supervisedly as a conditional model
        $\gprNo|\mappingNo$, see \fig\ref{fig:model_3}.
	The use of this feature space can reduce the complexity of the
        learning problem. For example, for complicated non-linear functions
        a higher-dimensional (overcomplete)
        representation~$\latentsSpace$ allows learning a simpler
        mapping \mbox{$\gprNo: \latentsSpace \to \targetsSpace$}. For
        high-dimensional inputs, the data often lies on a
        lower-dimensional manifold~$\latentsSpace$, e.g., due to
        non-discriminant or strongly correlated covariates. The
        lower-dimensional feature space~$\latentsSpace$ reduces the
        effect of the curse of dimensionality. In this paper, we focus
        on modeling complex functions with a relatively
        low-dimensional input space, which, nonetheless, cannot be well
        modeled by off-the-shelf GP {\gpkernel}s.
	
	    Typically, unsupervised feature learning methods determine the
        mapping~$\mappingNo$ by optimizing an unsupervised objective,
        independent from the objective of the overall
        regression~$\regressionNo$. Examples of such unsupervised
        objectives are the minimization of the input reconstruction
        error (auto-encoders~\citep{Vincent2008}), maximization of the
        variance (PCA~\citep{Pearson1901}), maximization of the
        statistical independence (ICA~\citep{Hyvarinen2000}), or the
        preservation of the distances between data
        (isomap~\citep{Tenenbaum2000} or LLE~\citep{Roweis2000}).
	In the context of regression, an unsupervised
        approach for feature learning can be insufficient as
	the learned data representation~$\latentsSpace$ might not
        suit the overall regression task~$\regressionNo$~\citep{Wahlstrom2015}:
        Unsupervised and supervised learning optimize different
        objectives, which do not necessarily match, e.g., minimizing
        the reconstruction error as unsupervised objective and
        maximizing the marginal likelihood as supervised objective.
	An approach where feature learning is performed in a
        supervised manner can instead guide learning the feature
        mapping~$\mappingNo$ toward representations that are useful
        for the overall regression~$\regressionNo = \gprNo \circ
        \mappingNo$. This intuition is the key insight of our
        {\newmethodlong}, where the feature mapping $\mappingNo$ and
        the GP $\gprNo$  are
        learned jointly using the same supervised objective as
        depicted in~\fig\ref{fig:model_4}.

\subsection{Gaussian Process Regression}

GPs are a state-of-the-art probabilistic non-parametric regression
method~\citep{Rasmussen2006}. Such a GP is a distribution over
functions
	\begin{align}
		\regressionNo \sim \GP \left( m,\coveq \right)
	\end{align}
	and fully defined by a mean function $m$ (in our case $m \equiv \vec 0$) and a {\gpkernel} $\coveq$.
	The GP predictive distribution at a test input $\vec x_*$ is given by
	\begin{align}
		\prob \left( \regression{\vec x_*}|\D,\vec x_* \right) &= \gauss{\mu(\vec x_*)}{\sigma^2(\vec x_*)}\,, 
		\label{eq:one-step prediction distr}\\
		\mu(\vec x_*) &= \vec k^T_*(\mat K + \sigma_w^2 \mat I)^{-1} \targets\,,
		\label{eq:one-step prediction mean}\\
		\sigma^2(\vec x_*) &= k_{**}-\vec k^T_*(\mat K + \sigma_w^2 \mat I)^{-1}\vec k_*\,,
	  	\label{eq:one-step prediction mean and covariance}
	\end{align}
	where $\D=\{\inputs,\targets\}$ is the training data, $\vec K$ is the kernel matrix with $K_{ij}= k(\vec x_i,\vec x_j)$, $k_{**}=k(\vec x_*,\vec x_*)$, \mbox{$\vec k_{*}=k(\inputs,\vec x_*)$} and $\sigma_w^2$ is the measurement noise variance. 	
	In our experiments, we use different {\gpkernel}s $\coveq$. Specifically, we use the squared exponential {\gpkernel} with Automatic Relevance Determination (ARD)
	\begin{align}
		\hspace{-7pt}{k_{\text{SE}}(\vec x_p,\vec x_q)} = \sigma_f^2\exp\left(-\tfrac{1}{2}(\vec x_p\! -\!\vec x_q)\T {\mat \Lambda\inv} (\vec x_p \!-\! \vec x_q)\right)\,,
		\label{eq:cov_SEARD}
	\end{align}
	with ${\mat \Lambda}=\diag([l^2_1,...,l^2_D])$, where $l_i$ are the characteristic length-scales, and $\sigma^2_f$ is the variance of the latent function~$\regressionNo$. Furthermore, we use the neural network {\gpkernel}
	\begin{align}
		\hspace{-5pt}{k_{\text{NN}}(\vec x_p,\vec x_q)} &= \sigma_f^2  \asin \left( \tfrac{\vec x_p\T \mat P \vec x_q}{\sqrt{(1+\vec x_p\T \mat P \vec x_p)(1+\vec x_q\T \mat P \vec x_q)}} \right)\,,
		\label{eq:cov_NN}
	\end{align}
	where $\mat P$ is a weight matrix.
	Each {\gpkernel} possesses various hyperparameters~$\parameters$ to be selected. This selection is performed by minimizing the Negative Log Marginal Likelihood (NLML)
		\begin{align}
			\text{NLML}(\parameters) &= - \log p(\targets | \inputs,\parameters) \label{eq:NLML_GP} \\&\stackrel{\cdot}{=} \tfrac{1}{2}\targets\T (\mat K_{\parameters} + \sigma_w^2 \mat I)\inv \targets  + \tfrac{1}{2} \log|\mat K_{\parameters} + \sigma_w^2 \mat I |\nonumber \,
            \end{align}
		%
		%
		Using the chain-rule, the corresponding gradient can be computed analytically as 
		\begin{align}
			\frac{\partial \text{NLML}(\parameters)}{\partial \parameters} &= 
			\color{red}\frac{\partial \text{NLML}(\parameters)}{\partial \mat K_{\parameters}} 
			\color{blue}\frac{\partial {\mat K}_{\parameters}}{\partial \parameters} \,,
		\end{align}
		which allows us to optimize the hyperparameters using Quasi-Newton optimization, e.g., \mbox{L-BFGS}~\citep{Liu1989}.


\subsection{\newmethodlong}
	In this section, we describe the {\newmethodshort} model and its parameters $\parameters_\text{\newmethodshort}$ itself, and relate it to standard GP regression. Furthermore, we detail training and prediction with the {\newmethodshort}.
	%
	\subsubsection{Model}
	 	As shown in \fig\ref{fig:model_4}, the {\newmethodshort} considers the overall regression as a composition of functions
		\begin{align}
			\regressionNo = \gprNo \circ \mappingNo \,.
			\label{eq:mgpdecomp}
		\end{align}
                The two functions $\mappingNo$ and $\gprNo$ are
                learned jointly to accomplish the overall regression
                objective function, i.e., the marginal likelihood in
                Equation~\eqref{eq:NLML_GP}. In this paper, we assume
                that $\mappingNo$ is a
                deterministic, parametrized function that maps the
                input space~$\inputsSpace$ into the feature space~$\latentsSpace \subseteq
                \R^{\dimLatents}$, which serves as the domain for the
                GP regression~$\gprNo: \latentsSpace \to
                \targetsSpace$. 
	Performing this transformation of the input data corresponds to training a GP~$\gprNo$ having $\latents =\mapping{\inputs}$ as inputs. Therefore, the {\newmethodshort} is equivalent to a GP for a function \mbox{$\regressionNo: \inputsSpace \to \targetsSpace$} with a {\gpkernel} $\tilde{k}$ defined as
	\begin{align}
		\tilde{k}(\vec x_p,\vec x_q) = k\left(\mapping{\vec x_p},\mapping{\vec x_q} \right) \,,
		\label{eq:kernel_mgp}
	\end{align}
	i.e., the kernel operates on the $\dimLatents$-dimensional feature space $\latentsSpace = \mapping{\inputsSpace}$.
	According to~\cite{MacKay1998}, a function defined as in \eq\eqref{eq:kernel_mgp} is a valid {\gpkernel} and, therefore, the {\newmethodshort} is a valid GP.
	
 		The predictive distribution for the {\newmethodshort} at a test input~$\vec x_*$ can then be derived from the predictive distribution of a standard GP in \eq\eqref{eq:one-step prediction distr}~as
	\begin{align}
		 \prob \left( \regression{\vec x_*}|\D,\vec x_* \right) &=  \prob \left( (\gprNo \circ \mappingNo) (\vec x_*)|\D,\vec x_* \right) \nonumber\\ 
		 								&=  \gauss{\mu(\mapping{\vec x_*})}{\sigma^2(\mapping{\vec x_*})}\,, 
		\label{eq:pred_mgp_1}\\
		 \mu(\mapping{\vec x_*}) &= \tilde{\vec k}^T_*(\tilde{\mat K} + \sigma_w^2 \mat I)^{-1} \targets\,,
		 \label{eq:pred_mgp_2}\\
 		 \sigma^2(\mapping{\vec x_*}) &= \tilde{k}_{**}-\tilde{\vec k}^T_*(\tilde{\mat K} + \sigma_w^2 \mat I)^{-1}\tilde{\vec k}_*\,,
	  	\label{eq:pred_mgp_3}
	\end{align}
	where $\tilde{\vec K}$ is the kernel matrix constructed as
        $\tilde{K}_{ij}= \tilde{k}(\vec x_i,\vec x_j)$,
        $\tilde{k}_{**}=\tilde{k}(\vec x_*,\vec x_*)$, $\tilde{\vec
          k}_{*}=\tilde{k}(\inputs,\vec x_*)$, and $\tilde{k}$ is the
        {\gpkernel} from \eq\eqref{eq:kernel_mgp}. In our experiments,
        we used the squared exponential {\gpkernel} from
        \eq\eqref{eq:cov_SEARD} for the kernel $k$ in
        \eq\eqref{eq:kernel_mgp}. %
	\subsubsection{Training}
	We train the {\newmethodshort} by jointly optimizing the parameters $\parameters_\mappingNo$ of the transformation~$\mappingNo$ and the GP hyperparameters~$\parameters_\gprNo$.
	For learning the parameters~$\parameters_\text{\newmethodshort} = [\parameters_\mappingNo, \parameters_\gprNo]$, we minimize the NLML as in the standard GP regression. Considering the composition of the mapping \mbox{$\regressionNo = \gprNo \circ \mappingNo$}, the NLML becomes
	\begin{align}
		&\text{NLML}(\parameters_\text{\newmethodshort}) = 
		- \log p \left( \targets | \inputs,\parameters_\text{\newmethodshort} \right) \nonumber \\
		&\stackrel{\cdot}{=} \tfrac{1}{2}\targets\T (\tilde{\mat K}_{\parameters_\text{\newmethodshort}} + \sigma_w^2 \mat I)\inv \targets \nonumber + \tfrac{1}{2} \log| \tilde{\mat K}_{\parameters_\text{\newmethodshort}} + \sigma_w^2 \mat I |.
		\label{eq:NLML_mGP}
	\end{align}
	Note that $\tilde{\mat K}_{\parameters_\text{\newmethodshort}}$ depends on both $\parameters_\gprNo$ and $\parameters_\mappingNo$, unlike $\mat K_{\parameters}$ from \eq\eqref{eq:NLML_GP}, which depends only on $\parameters_\gprNo$.
	The analytic gradients $\partial \text{NLML} / \partial \parameters_\gprNo$ of the objective in \eq\eqref{eq:NLML_mGP} with respect to the parameters $\parameters_\gprNo$ are computed as in the standard GP, i.e.,
	\begin{align}	
		\frac{\partial \text{NLML}(\parameters_\text{\newmethodshort}) }{\partial \parameters_\gprNo} &= 
		\color{red}\frac{\partial \text{NLML}(\parameters_\text{\newmethodshort}) }{\partial \mat K_{\parameters_\text{\newmethodshort}}}
		\color{blue}\frac{\partial \mat K_{\parameters_\text{\newmethodshort}}}{\partial \parameters_\gprNo} 
		\color{black}\,.
	\end{align}	
	 The gradients of the parameters~$\parameters_\mappingNo$ of the feature mapping are computed by applying the chain-rule
	\begin{align}
		\frac{\partial \text{NLML}(\parameters_\text{\newmethodshort}) }{\partial \parameters_\mappingNo} &=
		\color{red}\frac{\partial \text{NLML}(\parameters_\text{\newmethodshort}) }{\partial \mat K_{\parameters_\text{\newmethodshort}}}
		\color{black} \frac{\partial  \mat K_{\parameters_\text{\newmethodshort}}}{\partial \latents} 
		\color{darkgreen}\frac{\partial \latents}{\partial \parameters_\mappingNo}
		\color{black}\,,
		\label{eq:grad_mapping}
	\end{align}
	where only $\color{darkgreen} \partial \latents / \partial \parameters_\mappingNo$ depends on the chosen input transformation~$\mappingNo$, while $\partial  \mat K_{\parameters_\text{\newmethodshort}} / \partial \latents$ is the gradient of the kernel matrix with respect to the $\dimLatents$-dimensional GP training inputs~$\latents = \mapping{\inputs}$.
	%
	%
	Similarly to standard GP, the parameters~$\parameters_\text{\newmethodshort}$ in the {\newmethodshort} can be obtained using off-the-shelf optimization methods.
	%
%
\subsubsection{Input Transformation}
	Our approach can use any deterministic parametric data transformation $\mappingNo$. We focus on 
	%
	multi-layer neural networks and define their structure as
        $[\sizeLayer_1 - \dotsc - \sizeLayer_l]$ where $l$ is the
        number of layers, and $\sizeLayer_i$ is the number of neurons
        of the $i^\text{th}$ layer. Each layer $i=1,\dotsc,l$ of the
        neural network performs the transformation
	\begin{align}
		\nntransformation_i(\mat Z) &= \nntf{\nnW_i  \mat Z + \nnB_i}\,,
		\label{eq:NN}
	\end{align}
	where $\mat Z$ is the input of the layer, $\nntfNo$ is the
        transfer function, and $\nnW_i$ and $\nnB_i$ are the weights
        and the bias of the layer, respectively. Therefore, the input
        transformation~$\mappingNo$ of \eq\eqref{eq:mgpdecomp} is
        \mbox{$\mapping{\inputs} = (\nntransformation_l \circ\dotsc \circ
          \nntransformation_1)(\inputs)$}. 
 	The parameters~$\parameters_\mappingNo$ of the neural network
        $\mappingNo$ are the weights and biases of the whole network,
        so that $\parameters_\mappingNo = [\nnW_1,\nnB_1,\dotsc,
        \nnW_l,\nnB_l]$. The gradients $\color{darkgreen}\partial
        \latents /\partial \parameters_\mappingNo$ in
        \eq\eqref{eq:grad_mapping} are computed by repeated application of the chain-rule (backpropagation).
 	%


\section{Experimental Results}
\label{Results}

	\begin{figure*}[th]
		\centering
        \begin{subfigure}[t]{0.46\hsize}
        		\centering
                \includegraphics[width=\textwidth]{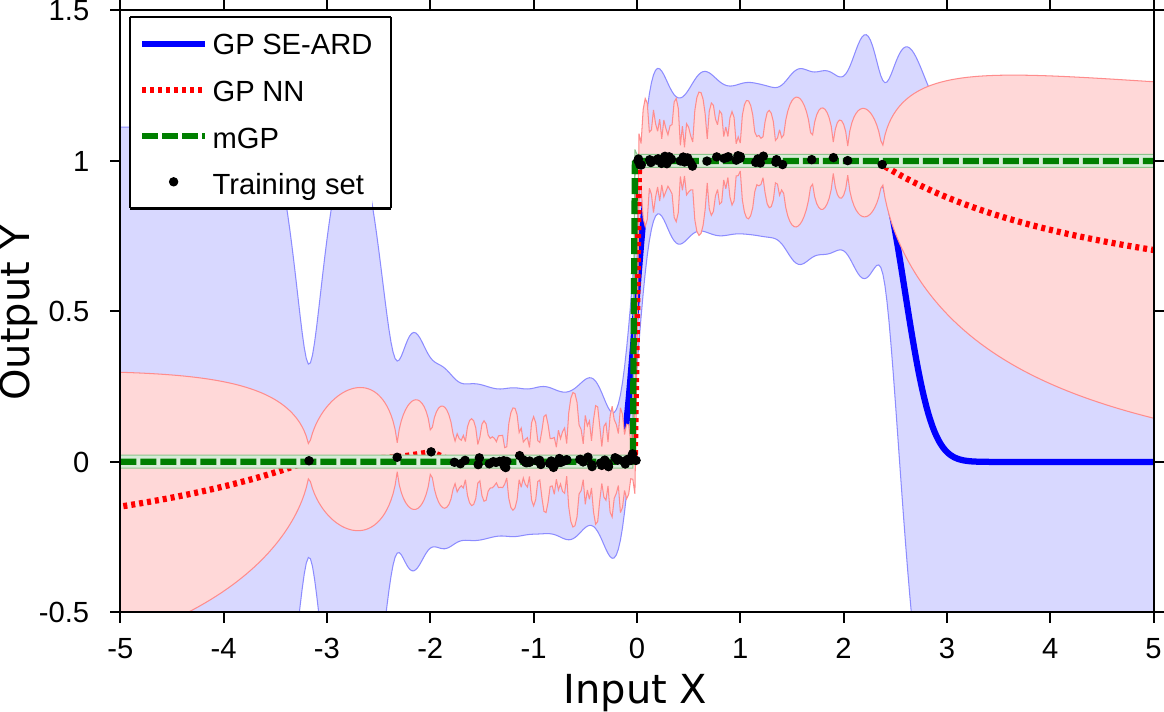}
                \caption{GP prediction.}
                \label{fig:discontinuous_1D_1a}
        \end{subfigure}%
        \hfill
        \begin{subfigure}[t]{0.46\hsize}
        	\centering
                \includegraphics[width=\textwidth]{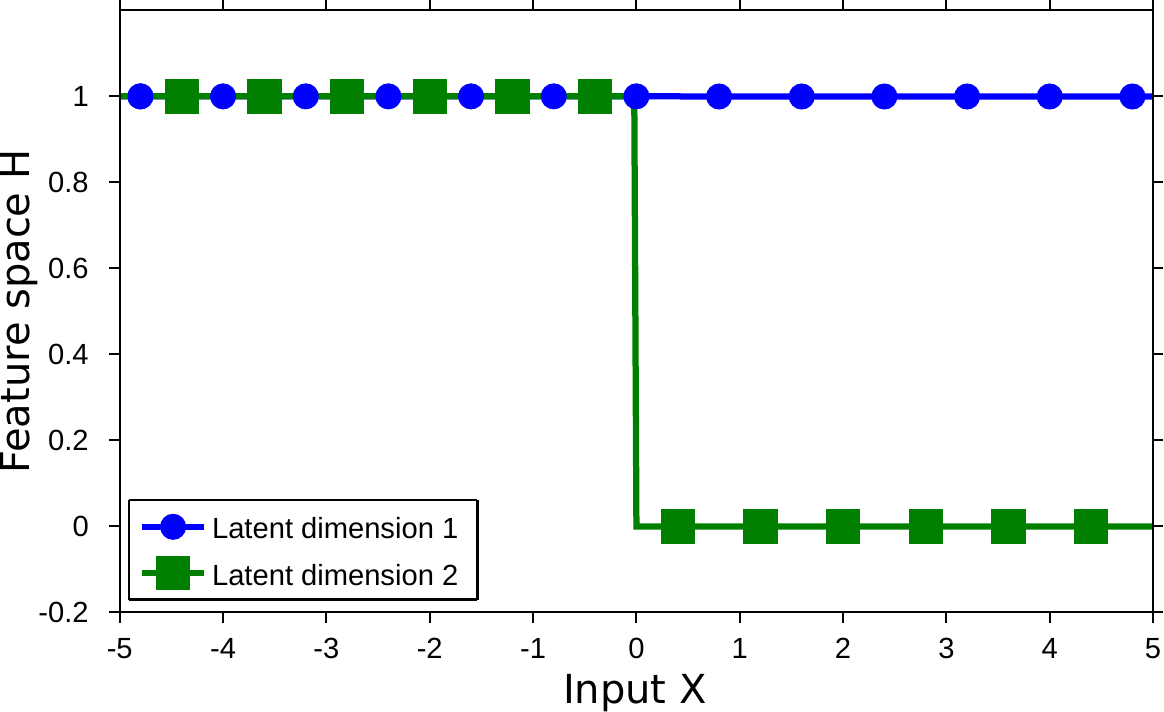}
                \caption{Learned mapping $\mappingNo$ using
                  {\newmethodshort} (log-sigmoid).}
                \label{fig:discontinuous_1D_1b}
        \end{subfigure}
        \caption{\textbf{\secExpOne}:
          (\subref{fig:discontinuous_1D_1a}) Predictive mean and 95\%
          confidence bounds for a GP with SE-ARD {\gpkernel}
          (\textcolor{blue}{blue solid}), a GP with NN {\gpkernel}
          (\textcolor{red}{red dotted}) and a log-sigmoid {\newmethodshort}
          (\textcolor{darkgreen}{green dashed}) on the step function
          of \eq\eqref{eq:discontinuous}. The discontinuity is
          captured better by an {\newmethodshort} than by a regular GP
          with either SE-ARD or NN
          {\gpkernel}s. (\subref{fig:discontinuous_1D_1b}) The 2D
          feature space~$\latentsSpace$ discovered by the non-linear
          mapping~$\mappingNo$ as a function of the
          input~$\inputsSpace$. The discontinuity of the modeled function
          is already captured by the non-linear
          mapping~$\mappingNo$. Hence, the mapping from feature
          space~$\latentsSpace$ to the output~$\targetsSpace$ is smooth and can
          be easily managed by the GP.}
		\label{fig:discontinuous_1D}
	\end{figure*}


        To demonstrate the efficiency of our proposed approach, we
        apply the mGP to challenging benchmark problems and a
        real-world regression task.
First, we demonstrate that {\newmethodshort}s can be successfully
applied to learning discontinuous functions, a daunting undertaking
with an off-the-shelf covariance function, due to its underlying
smoothness assumptions. 
Second, we evaluate {\newmethodshort}s on a function with multiple natural length-scales.
Third, we assess {\newmethodshort}s on real data from a walking
bipedal robot. The locomotion data set is highly challenging due to
ground contacts, which cause the regression function to violate
standard smoothness assumptions.  

To evaluate the goodness of the different models on the training set,
we consider the NLML previously introduced in \eq\eqref{eq:NLML_GP}
and \eqref{eq:NLML_mGP}. Additionally, for the test set, we make use
of the Negative Log Predictive Probability (NLPP) 
\begin{align}
	- \log \,p(\vec y = \vec y_* | \mat X, \vec x_*,\mat Y, \parameters)\,,
	\label{eq:NLPP}
\end{align}
where the $\vec y_*$ is the test target for the input $\vec x_*$ as computed for the standard GP in \eq\eqref{eq:one-step prediction distr} and \eqref{eq:pred_mgp_1} for the mGP model.

We compare our {\newmethodshort} approach with GPs using the SE-ARD
and NN covariance functions, which implement the model in
\fig\ref{fig:model_1}. Moreover, we evaluate two unsupervised feature
extraction methods, Random Embeddings and PCA, followed by a GP
SE-ARD, which implements the model in
\fig\ref{fig:model_3}.\footnote{The random embedding is computed as
  the transformation $\latents = \mat \alpha \inputs$, where the
  elements of $\mat \alpha$ are randomly sampled from a normal
  distribution.} For the model in \fig\ref{fig:model_4}, we consider
two variants of {\newmethodshort} with the log-sigmoid \mbox{$\nntf{x}
  = 1/(1+e^{-x})$} and the identity \mbox{$\nntf{x} = x$} transfer
functions. These two transfer functions lead to a
non-linear and a linear transformation $\mappingNo$, respectively.

%
%
%
%
%

\subsection{\secExpOne}
%

	In the following, we consider the step function
	%
	%
	%
	\begin{figure*}[t]
		\centering
        \begin{subfigure}[t]{0.32\textwidth}
        		\centering
                \includegraphics[width=\textwidth]{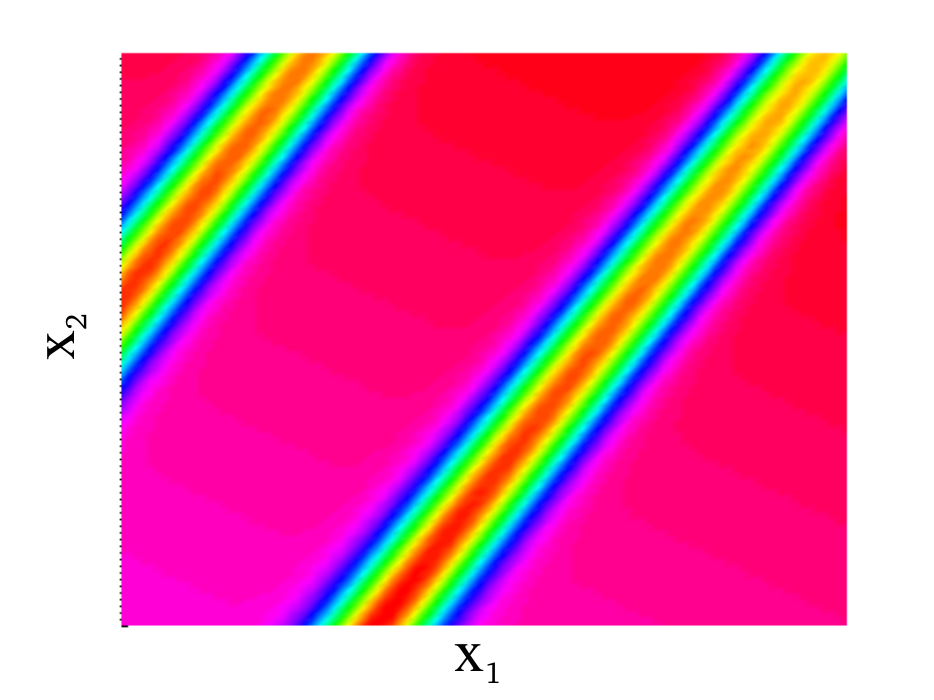}
                \caption{Intensity map of the function.}
                \label{fig:nasty_01}
        \end{subfigure}%
        \hfill
        \begin{subfigure}[t]{0.32\textwidth}
        		\centering
                \includegraphics[width=\textwidth]{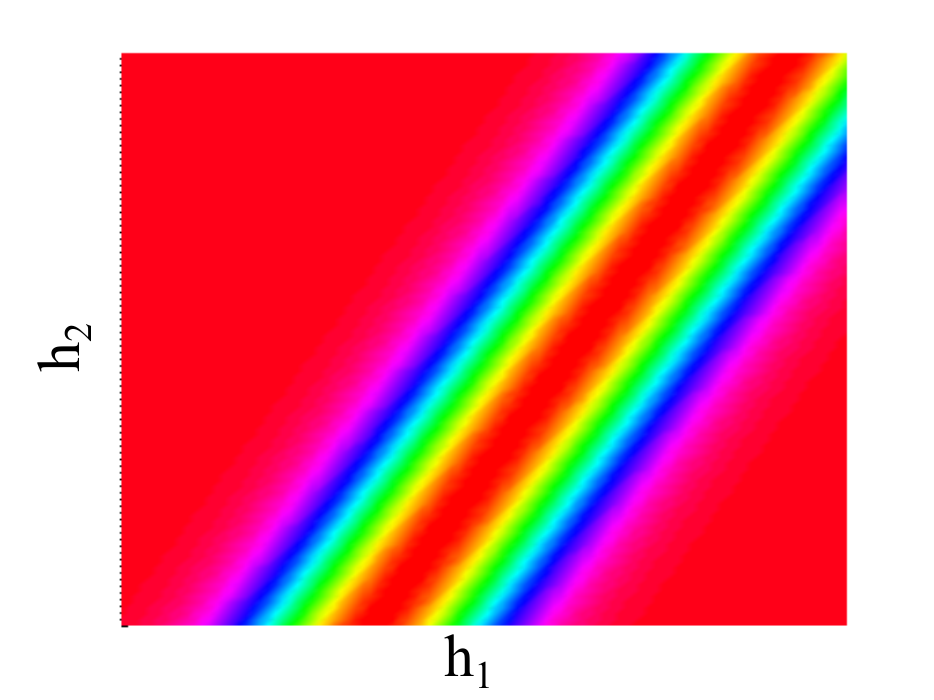}
                \caption{Intensity map of the learned feature space for the mGP (identity).}
                \label{fig:nasty_02}
        \end{subfigure}%
        \hfill
        \begin{subfigure}[t]{0.32\textwidth}
        		\centering
                \includegraphics[width=\textwidth]{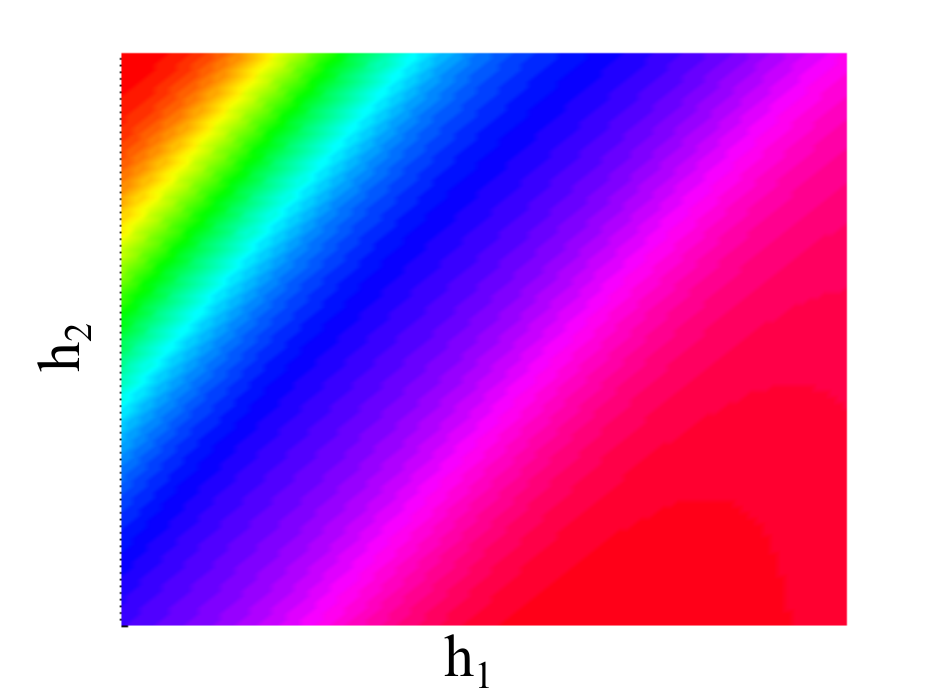}
                \caption{Intensity map of the learned feature space for the mGP (log-sigmoid).}
                \label{fig:nasty_03}
        \end{subfigure}%
        \\\vspace{5pt}
        \begin{subfigure}[t]{0.30\textwidth}
        		\centering
                \includegraphics[width=\textwidth]{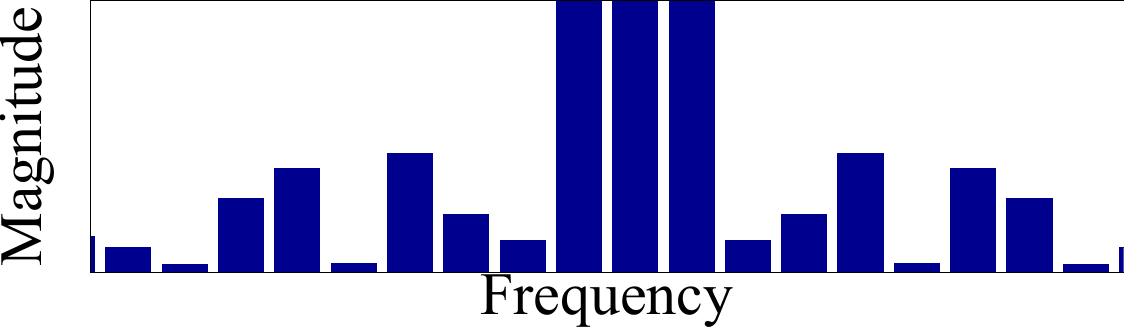}
                \caption{Spectrum of the function.}
                \label{fig:nasty_04}
        \end{subfigure}
        \hfill
        \begin{subfigure}[t]{0.30\textwidth}
        		\centering
                \includegraphics[width=\textwidth]{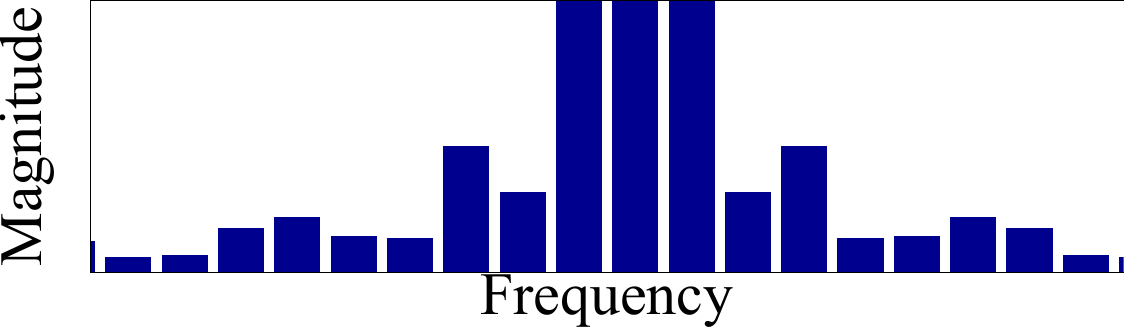}
                \caption{Spectrum of the learned feature space for the mGP (identity).}
                \label{fig:nasty_05}
        \end{subfigure}
        \hfill
        \begin{subfigure}[t]{0.30\textwidth}
        		\centering
                \includegraphics[width=\textwidth]{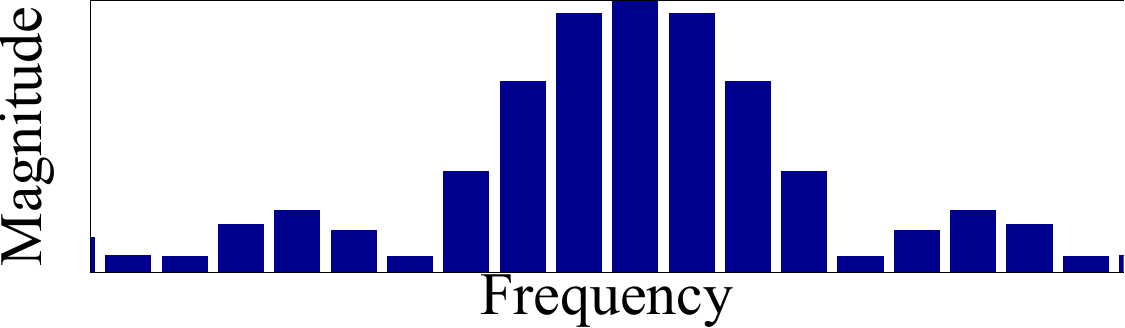}
                \caption{Spectrum of the learned feature space for the mGP (log-sigmoid).}
        \label{fig:nasty_06}
        \end{subfigure}
        \caption{\textbf{\secExpFour}: Intensity map of
          (\subref{fig:nasty_01}) the considered function,
          (\subref{fig:nasty_02}) the learned feature space of the mGP with a
          linear activation function and (\subref{fig:nasty_03}) with
          a log-sigmoid
          activation. (\subref{fig:nasty_04})--(\subref{fig:nasty_06})
          The corresponding Spectrum for
          (\subref{fig:nasty_04}) the original function and the
          learned feature space for (\subref{fig:nasty_05}) mGP
          (identity) and (\subref{fig:nasty_06}) mGP
          (log-sigmoid). The spectral analysis of the original
          function shows the presence of multiple frequencies. The
          transformations learned by both variants of mGP focus the
          spectrum of the feature space towards a more compact frequencies support.
}
		\label{fig:nasty}
	\end{figure*}
	\begin{align}
		y &= \regression{x} + w\,, \quad w \sim \gauss{0}{0.01^2}\,, \nonumber \\
 		\regression{x} &=
  		\begin{cases}
   			0 & \text{if } x \leq 0 \\
   			1 & \text{if } x > 0
  		\end{cases}\,.
  		\label{eq:discontinuous}
	\end{align}	
	For training, 100 inputs points are sampled from $\gauss{0}{1}$
        while the test set is composed of 500 data points uniformly
        distributed between $-5$ and $+5$. 
        The {\newmethodshort} uses a multi-layer neural network of
        [1-6-2] neurons (such that the feature space~$\latentsSpace
        \subseteq \R^{2}$) for the mapping~$\mappingNo$ and a standard
        \mbox{SE-ARD} {\gpkernel} for the GP
        regression~$\gprNo$. Values of the NLML per data point for the
        training and NLPP per data point for the test set are reported
        in \tab\ref{table:discontinuous_1D}. In both performance
        measures, the {\newmethodshort} using a non-linear
        transformation outperforms the other models.
	An example of the resulting predictive mean and the 95\%
        confidence bounds for three models is shown in
        \fig\ref{fig:discontinuous_1D_1a}.  Due to the implicit
        assumptions employed by the \mbox{SE-ARD} and NN {\gpkernel}s
        on the mapping~$\regressionNo$,
	neither of them appropriately captures the discontinuous
        nature of the underlying function or its correct noise level.
	The GP model applied to the random embedding and
        {\newmethodshort} (identity) perform similar to a standard GP
        with \mbox{SE-ARD} {\gpkernel} as their linear transformations
        do not substantially change the
        function. 
	Compared to these models, the {\newmethodshort} (log-sigmoid)
        captures the discontinuities of the function better, thanks to
        its non-linear transformation, while the uncertainty remains
        small over the whole function's domain.

	Note that the {\newmethodshort} still assumes smoothness in
        the regression~$\gprNo$, which requires the
        transformation~$\mappingNo$ to take care of the
        discontinuity. This effect can be observed in
        \fig\ref{fig:discontinuous_1D_1b}, where an example of the 2D
        learned feature space~$\latentsSpace$ is shown. The
        discontinuity is already encoded in the feature space. Hence,
        it is easier for the GP to learn the
        mapping~$\gprNo$. Learning the discontinuity in the feature
        space is a direct result from jointly training $\mappingNo$
        and $\gprNo$ as feature learning is embedded in the overall
        regression~$\regressionNo$.
\begin{table}[t]
		\centering
                \caption{\textbf{\secExpOne}: Negative Log Marginal Likelihood (NLML) and Negative Log Predictive Probability (NLPP) per data point for the step function of \eq\eqref{eq:discontinuous}. The {\newmethodshort} (log-sigmoid) captures the nature of the underlying function better than a standard GP in both the training and test sets.}
                \resizebox{\linewidth}{!} {
		\label{table:discontinuous_1D}
		\begin{tabular}{|l|c|c|c|c|}\hline
			\multicolumn{1}{|c|}{Method} 							& \multicolumn{2}{|c|}{Training set}			& \multicolumn{2}{|c|}{Test set} 			\\ 
			&NLML& RMSE & NLPP & RMSE \\ \hline
			GP SE-ARD 						& $-0.68 $	&	$ 1.00 \times 10^{-2} $	& $+0.50 \times 10^{-3}$ 		&$ 0.58$\\
			GP NN 							& $-1.49 $	&	$\mathbf{0.57 \times 10^{-2}  }$	& $+0.02 \times 10^{-3}$ 		& $0.14 $\\
			{\newmethodshort} (log-sigmoid)	& $\mathbf{-2.84}$& $ 1.06 \times 10^{-2}$	& $\mathbf{-6.34 \times 10^{-3}}$ 	& $\mathbf{0.02}$\\
			{\newmethodshort} (identity)	& $-0.68 $& 	$ 1.00 \times 10^{-2}$		& $+0.50 \times 10^{-3}$ 		& $ 0.58$\\
			RandEmb + GP SE-ARD				& $-0.77 $ &	$5.26 \times 10^{-2}	$	& $+0.51 \times 10^{-3}$	& $0.52$ \\ \hline
		\end{tabular}
        }
	\end{table}
	%


\subsection{\secExpFour}

In the following, we demonstrate that the {\newmethodshort} can be
used to model functions that possess multiple intrinsic length-scales. For
this purpose, we rotate the function
\begin{align}
\hspace{-5pt} y = 1 - \mathcal{N}\left( x_2|3,0.5^2 \right)- \mathcal{N}\left( x_2|-3,0.5^2 \right) + \tfrac{x_1}{100} 
	\label{eq:nasty}
\end{align}
	anti-clockwise by $45^\circ$. The intensity map of the resulting
	function is shown in \fig\ref{fig:nasty_01}. By itself (i.e., without
	rotating the function), \eq\eqref{eq:nasty} is a fairly simple
	function. However, when rotated, the correlation between the
	covariates substantially complicates modeling. If we
	consider a horizontal slice of the rotated function, we can see
	how different spectral frequencies are present in the function, see
	\fig\ref{fig:nasty_04}. The presence of different frequencies is
	problematic for {\gpkernel}s, such as the SE-ARD, which assume a
	single frequency. When learning the hyperparameters, the length-scale needs to
	trade off different frequencies. Typically, the hyperparameter optimization gives a preference to shorter
	length-scales. However, such a trade-off greatly reduces the generalization
	capabilities of the model.

        We compare the performances of a standard GP using SE-ARD and NN
        {\gpkernel}s and random embeddings followed by a
        GP using the SE-ARD {\gpkernel}, and our proposed {\newmethodshort}. We train
        these models with 400 data points, randomly sampled from a
        uniform distribution in the intervals $x_1 = [0,10]$, $x_2 =
        [0,10]$. As a test set we use 2500 data points distributed on
        a regular grid in the same intervals. For the
        {\newmethodshort} with both the log-sigmoid and the identify
        transfer functions, we use a neural network of [2-10-3]
        neurons. The NLML and the NLPP per data point are shown in \tab\ref{table:another}. The {\newmethodshort}
        outperforms all other methods evaluated. We believe that
        this is due to the mapping~$\mappingNo$, which transforms the
        input space so as to have a single natural frequency.
        \fig\ref{fig:nasty_02} shows the intensity map of the feature
        space after the {\newmethodshort} transformed the inputs using
        a neural network with the identify transfer function.
        \fig\ref{fig:nasty_03} shows the intensity map of the feature
        when the log-sigmoid transfer function is used. Both transformations tend to make the feature
        space smother compared to the initial input space. This effect
        is the result of the transformations, which aim to equalize the
        natural frequencies of the original function in order to
        capture them more efficiently with a single length-scale. The
        effects of these transformations are clearly visible in the
        spectrogram of the {\newmethodshort} (identity) in
        \fig\ref{fig:nasty_05} and of the {\newmethodshort}
        (log-sigmoid) in \fig\ref{fig:nasty_06}.
	The smaller support of the spectrum, obtained through the non-linear transformations performed by mGP using the log-sigmoid transfer function, translates into superior prediction performance.

	\begin{table}[t]
		\centering
		\caption{\textbf{\secExpFour}: NLML per data point for the training set and NLPP per data point for the test set. The
          {\newmethodshort} captures the nature of the underlying
          function better than a standard GP in both the training and
          test sets.} 
	\resizebox{\linewidth}{!} {
		\label{table:another}
		\begin{tabular}{|l|c|c|c|c|}\hline
			\multicolumn{1}{|c|}{Method} 							& \multicolumn{2}{|c|}{Training set}			& \multicolumn{2}{|c|}{Test set} 			\\ 
			&NLML& RMSE & NLPP & RMSE \\ \hline
			GP SE-ARD 					& $-2.46 $	& $0.40 \times 10^{-3} $		& $-4.34$ 		& $1.51\times 10^{-2}$\\
			GP NN 						& $-1.57 $	& $1.52 \times 10^{-3}$			& $-2.53$ 		& $6.32\times 10^{-2}$\\
			{\newmethodshort} (log-sigmoid)			& $\mathbf{-6.61}$& $\mathbf{ 0.37 \times 10^{-4}}$	& $\mathbf{-7.37}$ 	& $\mathbf{ 0.58 \times 10^{-4}}$\\
			{\newmethodshort} (identity)			& $-5.60$ 	& $ 0.79 \times 10^{-4}$		& $-6.63$ 		& $2.36\times 10^{-3}$\\
			RandEmb + GP SE-ARD				& $-0.47$ 	& $6.84 \times 10^{-3}$			& $-1.29$		& $1.19\times 10^{-1}$ \\ \hline
		\end{tabular}
        }
	\end{table}
	%

%
\subsection{\secExpTwo}

\begin{wrapfigure}{r}{0.54\hsize}
        	\centering
                \includegraphics[width = 0.98\hsize]{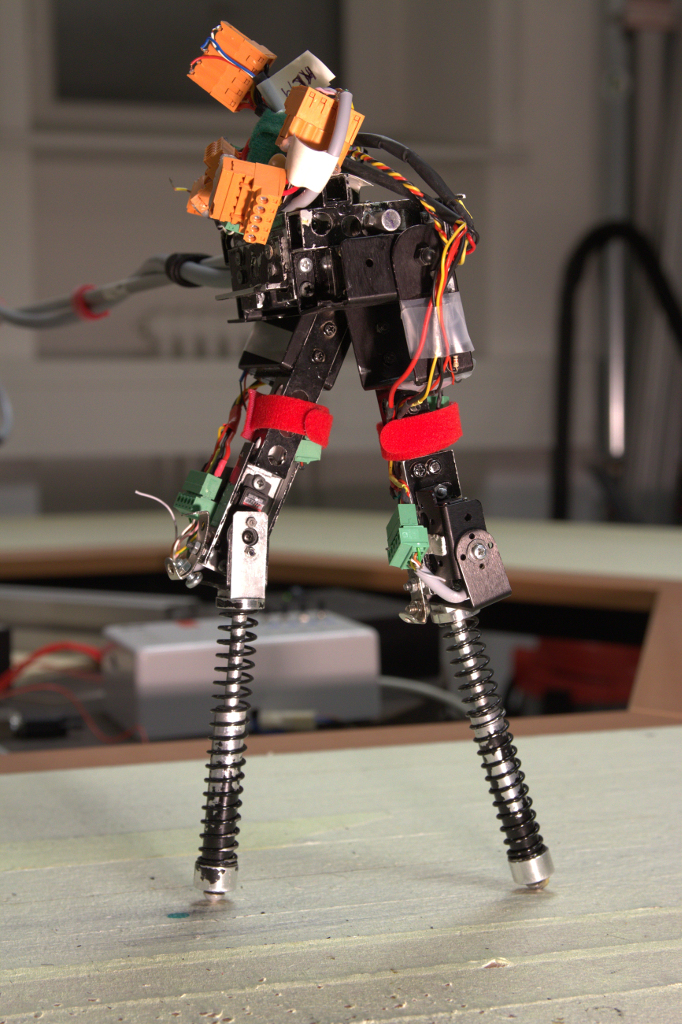}
                \caption{\textbf{\secExpTwo}: The bio-inspired bipedal walker \textit{Fox} from which the dataset is generated.}
                \label{fig:fox_fox}
\end{wrapfigure}

Modeling data from real robots can be challenging when the robot has
physical interactions with the environment. Especially in bipedal
locomotion, we lack good contact force and friction models. Thus, we
evaluate our {\newmethodshort} approach on modeling data from the
bio-inspired bipedal walker~\textit{Fox} \citep{Renjewski2012} shown in
\fig\ref{fig:fox_fox}. The data set consists of measurements of six
covariates recorded at regular intervals of $0.0125$\,sec. The
covariates are the angles of the right and left hip joints, the angles
of the right and left knee joints and two binary signals from the
ground contact sensors. We consider the regression task where the left
knee joint is the prediction target~$\targetsSpace$ and the remaining
five covariates are the inputs~$\inputsSpace$.  For training we
extract 400 consecutive data points, while we test on the following
500 data points. The {\newmethodshort} uses a network structure
[1-30-3].

\tab\ref{table:fox} shows that the {\newmethodshort} models the data
better than the other models. The standard GPs with SE-ARD or NN
{\gpkernel} predict the knee angle relatively
well. 

\begin{figure*}[t]
		\centering
                \begin{subfigure}[t]{0.32\hsize}
        		\centering
                \includegraphics[width = \hsize]{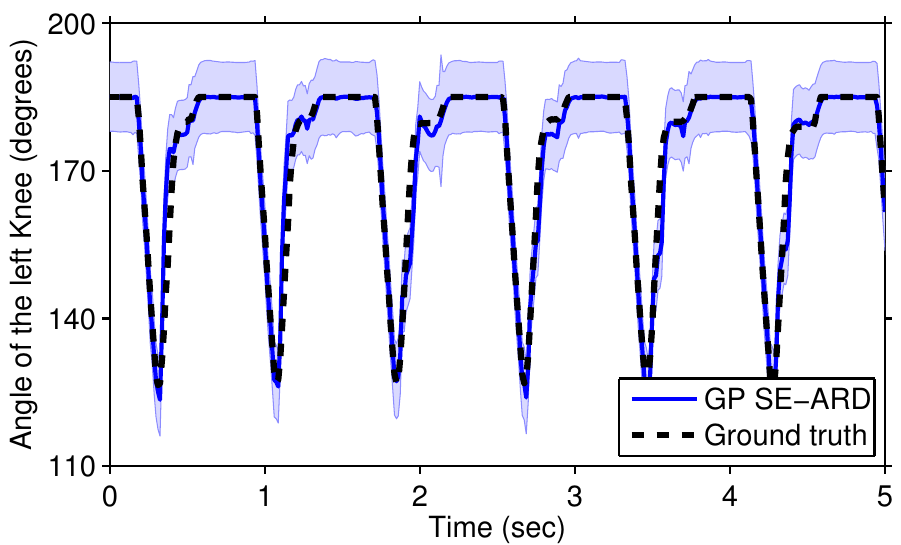}
                \caption{GP with SE-ARD}
                \label{fig:fox_pred_1}
        \end{subfigure}
         \hfill
                \begin{subfigure}[t]{0.32\hsize}
        		\centering
                \includegraphics[width = \hsize]{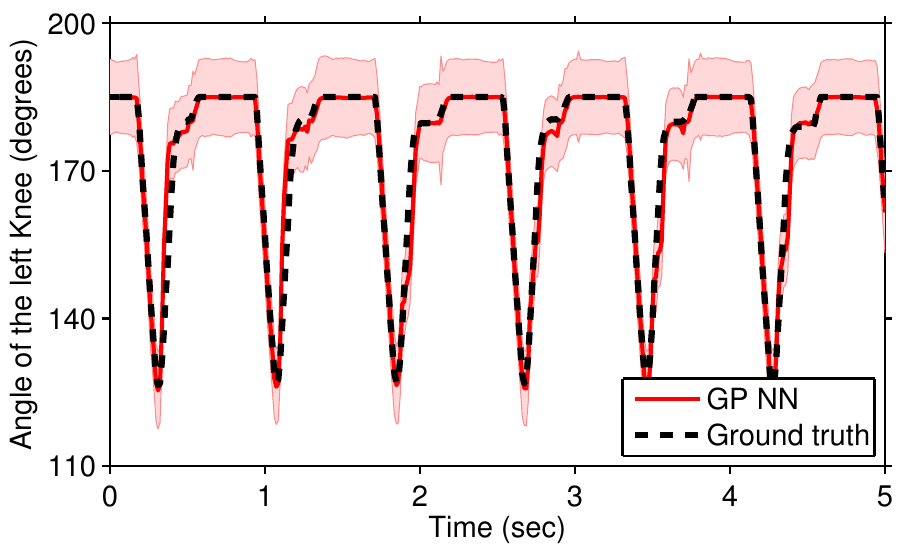}
                \caption{GP with NN}
                \label{fig:fox_pred_2}
        \end{subfigure}
                \hfill
                \begin{subfigure}[t]{0.32\hsize}
        		\centering
                \includegraphics[width = \hsize]{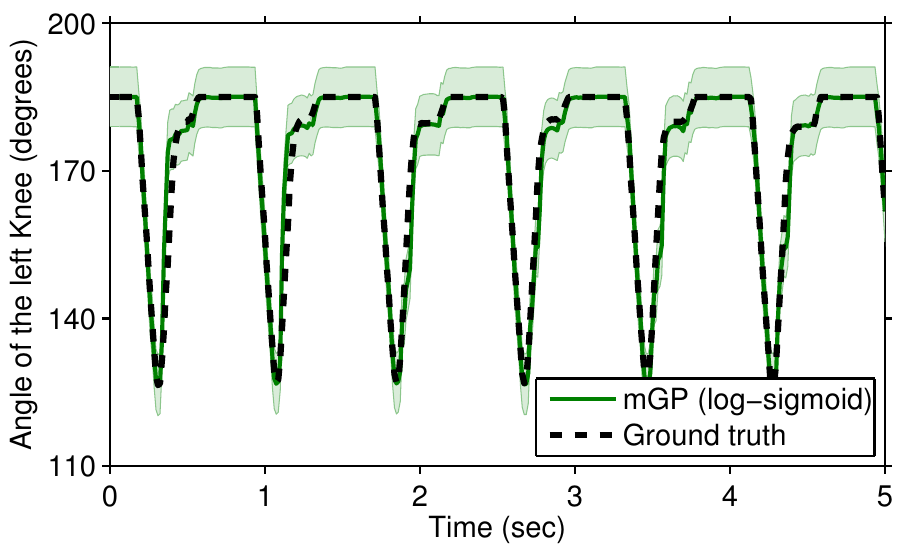}
                \caption{{\newmethodshort} (log-sigmoid) }
                \label{fig:fox_pred_3}
        \end{subfigure}
        \caption{\textbf{\secExpTwo}: Predictive mean and 95\%
          confidence bounds on the test set on real robot walking data
          for (\subref{fig:fox_pred_1}) GP with SE-ARD {\gpkernel},
          (\subref{fig:fox_pred_2}) GP with NN {\gpkernel} and
          (\subref{fig:fox_pred_3}) {\newmethodshort}
          (log-sigmoid). The {\newmethodshort} (log-sigmoid) captures
          the structure of the data better compared to GPs with either
          SE-ARD or NN {\gpkernel}s.}
        \label{fig:fox_pred}
	\end{figure*}

\begin{figure}[t]
		\centering
        		\centering
                \includegraphics[width = 0.8\hsize]{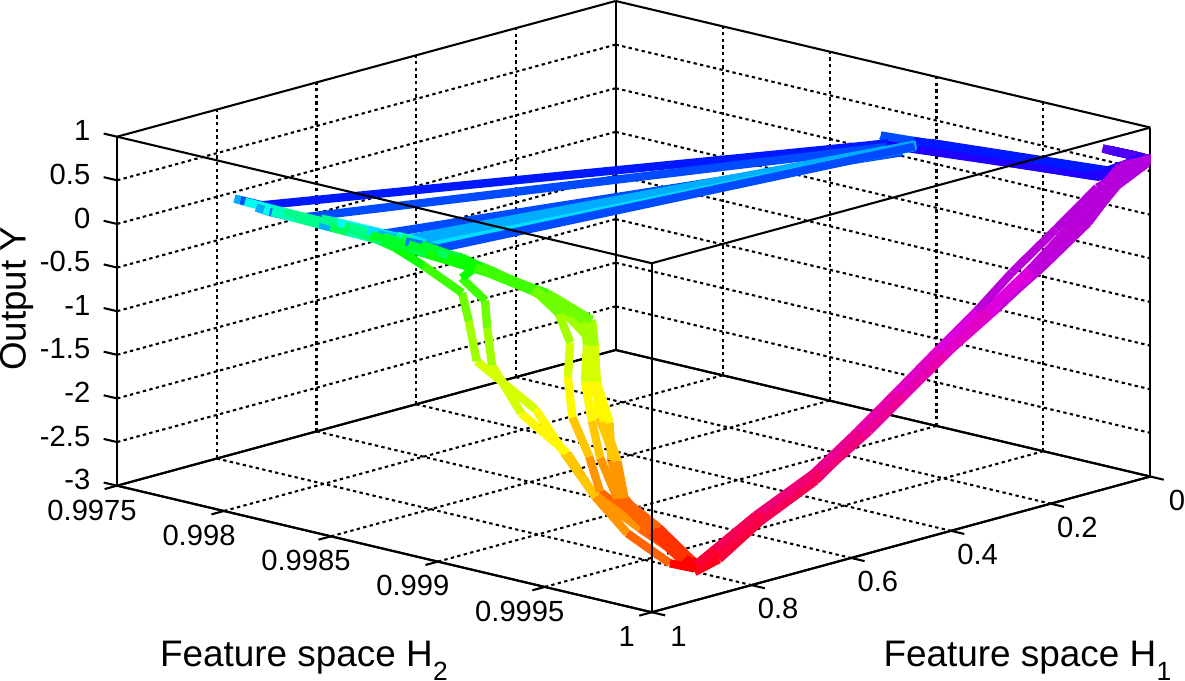}
                \caption{\textbf{\secExpTwo}: The mGP learns a smooth
                  feature space representation (the color indicates the phase during a single step). }
                \label{fig:fox_feature_1}
	\end{figure}

\fig\ref{fig:fox_pred} shows that the {\newmethodshort} has
larger variance of the prediction for areas where fast movement occurs
due to leg swinging. However, it captures the structure and regularity
of the data better, such as the mechanically 
enforced upper bound at
185 degrees. The uncertainty of about $20$ degrees is reasonable for
the fast changes in the knee angle during the swinging phase. However,
the same uncertainty of noise is unrealistic once the knee is fully
extended at 185 degrees.  Therefore, for control purposes, using the
{\newmethodshort} model would be preferable. This is a positive sign
of the potential of {\newmethodshort} to learn representations that
are meaningful for the overall regression task.
Figure~\ref{fig:fox_feature_1} visualizes two dimensions of the
learned feature space in which the walking trajectory is
smoothly embedded.

	\begin{table}[t]
	\centering
	\caption{\textbf{\secExpTwo}: NLML per data point for the
          training set and NLPP per data point for the test set. The
          {\newmethodshort} captures the nature of the underlying
          function well in both the training and test sets.} 
	\label{table:fox}
	\resizebox{\linewidth}{!} {
\begin{tabular}{|l|c|c|c|c|}\hline
			\multicolumn{1}{|c|}{Method} 							& \multicolumn{2}{|c|}{Training set}			& \multicolumn{2}{|c|}{Test set} 			\\ 
			&NLML& RMSE & NLPP & RMSE \\ \hline
            		GP SE-ARD 			& $-0.01 	$	&	$ 0.18$& $-0.13$ & $0.20$\\
            		GP NN 				& $ 0.04	$	&$\mathbf{0.17}$& $-0.13$& $0.20$\\
					{\newmethodshort} (log-sigmoid)	& $\mathbf{-0.28}$ 	& $\mathbf{0.17}$& $\mathbf{-0.18}$ &	$\mathbf{0.19}$\\
					{\newmethodshort} (identity)	& $0.97$ 	&	$0.03$& $0.86$ &		$0.66$\\
            		PCA + GP SE-ARD		& $0.01$ & $0.18$  & $-0.12$ & $0.20$	\\
            		RandEmb + GP SE-ARD	& $0.16$	&	$0.18$	& $-0.09$ & $0.20$ \\\hline
		\end{tabular}
        }
	\end{table}


\section{Discussion}
\label{discussion}


Unlike neural networks, which have been successfully used to extract
complex features, \cite{MacKay1998} argued that GPs are unsuited for feature
learning. However, with growing complexity of the regression problem,
the discovery of useful data representations (i.e., features) is often
necessary.
Despite similarities between neural networks and
GPs~\citep{Neal1995}, it is still unclear how to exploit the best of
both worlds.
Inspired by deep neural networks, deep GPs stack multiple layers of GP latent variable models~\citep{Damianou2013a}. 
This method can be used also in a supervised regression framework, which renders it similar to our proposed mGP. 
The main differences between Deep GPs and the mGP is that (a) Deep GPs integrate out the latent functions connecting the individual layers and do not require to explicitly define a deterministic transformation structure and, (b) unlike the mGP, the Deep GP is not a full GP.
Our {\newmethodshort} model extends a standard GP by learning corresponding useful data representations for the regression problem at hand. 
In particular, it can discover feature representations that comply with the implicit assumptions of the GP covariance function employed. 

One of the main challenges of training {\newmethodshort}s using neural
networks as mapping~$\mappingNo$ is the unwieldy joint optimization of
the parameters $\parameters_\text{\newmethodshort}$. The difficulty
resides in the non-convexity of the NLML, leading to multiple local
optima. Depending on the number of parameters $\parameters_\mappingNo$
of the feature map $\mappingNo$, the problem of local optima can be
severe. This problem is well-known for neural networks, and there may
be feasible alternatives to L-BFGS, such as the Hessian-free
optimization proposed by~\cite{Martens2010}. Additionally, sparsity
and low-rank approximations in $\nnW$ and $\nnB$ can be beneficial to
reduce the complexity of the optimization.

The extreme expressiveness of the mGP does not prevent the model from
solving ``easy'' regression tasks. For a proof-of-concept, we applied
the mGP to modeling a sinusoidal function, which is very easy to model
with a standard GP.  The results in \tab\ref{table:sin} suggest that
even for simple functions the mGP performs as good as a standard GP.

Increasing the number of parameters of the mapping~$\mappingNo$ intuitively leads to an increased flexibility in the learned {\gpkernel}. 
However, when the number of parameters exceeds the size of data set, the model is prone to over-fitting.  
For example, during experimental validation of the step function, we noticed the undesirable effect that the {\newmethodshort} could model discontinuities at locations slightly offset from their actual locations. 
In these cases, training data was sparse around the locations of discontinuity.  
This observed effect is due to over-fitting of the deterministic transformation~$\mappingNo$. 
Ideally, we replace this deterministic mapping with a probabilistic one, which would describe the uncertainty about the location of the discontinuity. 
In a fully Bayesian framework, we would average over possible models of the discontinuity. 
However, the use of such a probabilistic mapping in the context of GP regression is analytically intractable in closed form~\citep{Schmidt2003} and would require to train GPs with uncertain inputs. 
This kind of GP training is also analytically intractable, although approximations exist~\citep{Lawrence2005,Wang2008,McHutchon2011,Titsias2010}.

 	\begin{table}[t]
 	\centering
 	\caption{\textbf{Smooth function}: NLML per data point for the training set and NLPP per data point for the test set. There is no relevant difference between mGP and standard GP in modeling smooth functions.}
 	\label{table:sin}
 	\resizebox{\linewidth}{!} {
 \begin{tabular}{|l|c|c|c|c|}\hline
 			\multicolumn{1}{|c|}{Method} 	 & \multicolumn{2}{|c|}{Training set}	 & \multicolumn{2}{|c|}{Test set} \\ 
 			&NLML& RMSE & NLPP & RMSE \\ \hline
             		GP SE-ARD 			& $-4.30$	& $2.78\times 10^{-3}$ & $-4.42$ & $2.91\times 10^{-3}$\\
 			{\newmethodshort} (log-sigmoid)& $-4.31$ 	& $2.76\times 10^{-3}$ & $-4.42$ & $2.90\times 10^{-3}$\\
\hline
 		\end{tabular}
         }
 	\end{table}
	%


\section{Conclusion}
\label{conclusions}



The quality of a Gaussian process model  strongly depends on an appropriate covariance function. However,
 designing such a covariance function is challenging for some classes of
 functions, e.g., highly non-linear functions.
%
%
%
 To model such complex functions we introduced {\newmethodlong}. The
 key idea is to decompose the overall regression into learning a
 feature space mapping and a GP regression that maps from this feature
 space to the observed space. Both the input transformation and the GP
 regression are learned jointly and supervisedly by maximizing the
 marginal likelihood. The {\newmethodshort} is a valid
 GP for the overall regression task using a more expressive
 {\gpkernel}.

The {\newmethodshort} successfully modeled highly non-liner functions,
e.g., step functions or effects of ground contacts in robot locomotion, where standard GPs fail.
Applications that profit from the enhanced modeling
capabilities of the {\newmethodshort} include robot modeling (e.g., contact and stiction modeling), reinforcement learning, and Bayesian optimization. 


\ifCLASSOPTIONcompsoc
  \section*{Acknowledgments}
\else
  \section*{Acknowledgment}
\fi


The research leading to these results has received funding from the European Council under grant agreement \#600716 (CoDyCo - FP7/2007--2013). 
M.~P. Deisenroth was supported by a Google Faculty Research Award.




\bibliographystyle{abbrvnat}
\bibliography{ijcnn2016}

\begin{thebibliography}{26}
\providecommand{\natexlab}[1]{#1}
\providecommand{\url}[1]{\texttt{#1}}
\expandafter\ifx\csname urlstyle\endcsname\relax
  \providecommand{\doi}[1]{doi: #1}\else
  \providecommand{\doi}{doi: \begingroup \urlstyle{rm}\Url}\fi

\bibitem[Damianou and Lawrence(2013)]{Damianou2013a}
A.~C. Damianou and N.~D. Lawrence.
\newblock {Deep Gaussian Processes}.
\newblock In \emph{AISTATS}, 2013.

\bibitem[Duvenaud et~al.(2013)Duvenaud, Lloyd, Grosse, Tenenbaum, and
  Ghahramani]{Duvenaud2013a}
D.~Duvenaud, J.~R. Lloyd, R.~Grosse, J.~B. Tenenbaum, and Z.~Ghahramani.
\newblock {Structure Discovery in Nonparametric Regression through
  Compositional Kernel Search}.
\newblock In \emph{ICML}, 2013.

\bibitem[HajiGhassemi and Deisenroth(2014)]{HajiGhassemi2014}
N.~HajiGhassemi and M.~P. Deisenroth.
\newblock Approximate {Inference for Long-Term Forecasting with Periodic
  Gaussian Processes}.
\newblock In \emph{AISTATS}, 2014.

\bibitem[Hyv{\"a}rinen and Oja(2000)]{Hyvarinen2000}
A.~Hyv{\"a}rinen and E.~Oja.
\newblock {Independent Component Analysis: Algorithms and Applications}.
\newblock \emph{Neural Networks}, 13\penalty0 (4):\penalty0 411--430, 2000.

\bibitem[Lawrence(2005)]{Lawrence2005}
N.~D. Lawrence.
\newblock Probabilistic {N}on-linear {P}rincipal {C}omponent {A}nalysis with
  {G}aussian {P}rocess {L}atent {V}ariable {M}odels.
\newblock \emph{JMLR}, 6:\penalty0 1783--1816, November 2005.

\bibitem[Liu and Nocedal(1989)]{Liu1989}
D.~C. Liu and J.~Nocedal.
\newblock {On the Limited Memory BFGS Method for Large Scale Optimization}.
\newblock \emph{Mathematical Programming}, 45\penalty0 (3):\penalty0 503--528,
  1989.

\bibitem[MacKay(1998)]{MacKay1998}
D.~J.~C. MacKay.
\newblock Introduction to {Gaussian Processes}.
\newblock In \emph{Neural {Networks and Machine Learning}}, volume 168, pages
  133--165. 1998.

\bibitem[Martens(2010)]{Martens2010}
J.~Martens.
\newblock Deep {Learning via Hessian-free Optimization}.
\newblock In \emph{ICML}, 2010.

\bibitem[McHutchon and Rasmussen(2011)]{McHutchon2011}
A.~McHutchon and C.~E. Rasmussen.
\newblock Gaussian {Process Training with Input Noise}.
\newblock In \emph{NIPS}, 2011.

\bibitem[Neal(1995)]{Neal1995}
R.~M. Neal.
\newblock \emph{Bayesian {Learning for Neural Networks}}.
\newblock PhD thesis, University of Toronto, 1995.

\bibitem[Pearson(1901)]{Pearson1901}
K.~Pearson.
\newblock On {Lines and Planes of Closest Fit to Systems of Points in Space}.
\newblock \emph{The London, Edinburgh, and Dublin Philosophical Magazine and
  Journal of Science}, 2\penalty0 (11):\penalty0 559--572, 1901.

\bibitem[Rasmussen and Williams(2006)]{Rasmussen2006}
C.~E. Rasmussen and C.~K.~I. Williams.
\newblock \emph{Gaussian {Processes for Machine Learning}}.
\newblock The MIT Press, 2006.

\bibitem[Renjewski(2012)]{Renjewski2012}
D.~Renjewski.
\newblock \emph{An {Engineering Contribution to Human Gait Biomechanics}}.
\newblock PhD thesis, TU Ilmenau, 2012.

\bibitem[Roweis and Saul(2000)]{Roweis2000}
S.~T. Roweis and L.~K. Saul.
\newblock Nonlinear {Dimensionality Reduction by Locally Linear Embedding}.
\newblock \emph{Science}, 290\penalty0 (5500):\penalty0 2323--2326, 2000.

\bibitem[Salakhutdinov and Hinton(2007)]{Salakhutdinov2007a}
R.~Salakhutdinov and G.~Hinton.
\newblock Using {Deep Belief Nets to Learn Covariance Kernels for Gaussian
  Processes}.
\newblock In \emph{NIPS}, 2007.

\bibitem[Schmidt and O'Hagan(2003)]{Schmidt2003}
A.~M. Schmidt and A.~O'Hagan.
\newblock Bayesian {Inference for Non-stationary Spatial Covariance Structure
  via Spatial Deformations}.
\newblock \emph{Journal of the Royal Statistical Society: Series B (Statistical
  Methodology)}, 65\penalty0 (3):\penalty0 743--758, 2003.

\bibitem[Snelson and Ghahramani(2006)]{Snelson2006}
E.~Snelson and Z.~Ghahramani.
\newblock {Variable Noise and Dimensionality Reduction for Sparse Gaussian
  Processes}.
\newblock In \emph{UAI}, 2006.

\bibitem[Snelson et~al.(2004)Snelson, Rasmussen, and Ghahramani]{Snelson2004}
E.~Snelson, C.~E. Rasmussen, and Z.~Ghahramani.
\newblock {Warped Gaussian Processes}.
\newblock In \emph{NIPS}, 2004.

\bibitem[Snoek et~al.(2012)Snoek, Adams, and Larochelle]{Snoek2012b}
J.~Snoek, R.~P. Adams, and H.~Larochelle.
\newblock Nonparametric {Guidance of Autoencoder Representations using Label
  Information}.
\newblock \emph{JMLR}, 13:\penalty0 2567--2588, 2012.

\bibitem[Snoek et~al.(2014)Snoek, Swersky, Zemel, and Adams]{Snoek2014a}
J.~Snoek, K.~Swersky, R.~S. Zemel, and R.~P. Adams.
\newblock {Input Warping for Bayesian Optimization of Non-stationary
  Functions}.
\newblock In \emph{ICML}, 2014.

\bibitem[Tenenbaum et~al.(2000)Tenenbaum, De~Silva, and
  Langford]{Tenenbaum2000}
J.~B. Tenenbaum, V.~De~Silva, and J.~C. Langford.
\newblock A {Global Geometric Framework for Nonlinear Dimensionality
  Reduction}.
\newblock \emph{Science}, 290\penalty0 (5500):\penalty0 2319--2323, 2000.

\bibitem[Titsias and Lawrence(2010)]{Titsias2010}
M.~K. Titsias and N.~D. Lawrence.
\newblock {Bayesian Gaussian Process Latent Variable Model}.
\newblock In \emph{AISTATS}, 2010.

\bibitem[Vincent et~al.(2008)Vincent, Larochelle, Bengio, and
  Manzagol]{Vincent2008}
P.~Vincent, H.~Larochelle, Y.~Bengio, and P.-A. Manzagol.
\newblock Extracting and {Composing Robust Features with Denoising
  Autoencoders}.
\newblock \emph{ICML}, 2008.

\bibitem[Wahlstr\"om et~al.(2015)Wahlstr\"om, Sch\"on, and
  Deisenroth]{Wahlstrom2015}
N.~Wahlstr\"om, T.~B. Sch\"on, and M.~P. Deisenroth.
\newblock Learning {Deep Dynamical Models From Image Pixels}.
\newblock In \emph{SYSID}, 2015.

\bibitem[Wang et~al.(2008)Wang, Fleet, and Hertzmann]{Wang2008}
J.~M. Wang, D.~J. Fleet, and A.~Hertzmann.
\newblock Gaussian {Process Dynamical Models for Human Motion}.
\newblock \emph{PAMI}, 30\penalty0 (2):\penalty0 283--298, 2008.

\bibitem[Wilson and Adams(2013)]{Wilson2013}
A.~G. Wilson and R.~P. Adams.
\newblock Gaussian {Process Kernels for Pattern Discovery and Extrapolation}.
\newblock \emph{ICML}, 2013.

\end{thebibliography}

\end{document}